\documentclass[10pt,twocolumn,letterpaper]{article}

\usepackage{cvpr}
\usepackage{times}
\usepackage{epsfig}
\usepackage{graphicx}
\usepackage{amsmath}
\usepackage{amssymb}
\usepackage{indentfirst}

\usepackage{fancyref}
\usepackage{float}

\usepackage[breaklinks=true,bookmarks=false,hidelinks]{hyperref}

\cvprfinalcopy{} 
\def\cvprPaperID{****} 

\newcommand*{\idiaereses}{\"\i}
\newcommand*{\naive}{na\idiaereses{}ve}
\newcommand*{\Naive}{Na\idiaereses{}ve}

\numberwithin{equation}{section}

\begin{document}

\title{Learning to Predict the 3D Layout of a Scene}

\author{Jihao Andreas Lin \qquad
    Jakob Br{\"u}nker \qquad
    Daniel F{\"a}hrmann \\ \\
    Department of Computer Science, TU Darmstadt
}
\maketitle

\begin{abstract}

While 2D object detection has improved significantly over the past
years~\cite{RCNN, FastRCNN, FasterRCNN, MaskRCNN, YOLO, RetinaNet}, real world
applications of computer vision often require an understanding of the 3D layout
of a scene. Many recent approaches to 3D detection~\cite{PointFusion, AVOD,
VoxelNet, FrustumPointNet} use LiDAR point clouds for prediction. We propose a
method that only uses a single RGB image, thus enabling applications in devices
or vehicles that do not have LiDAR sensors. By using an RGB image, we can
leverage the maturity and success of recent 2D object detectors, by extending a
2D detector with a 3D detection head. In this paper we discuss different
approaches and experiments, including both regression and classification
methods, for designing this 3D detection head. Furthermore, we evaluate how
subproblems and implementation details impact the overall prediction result. We
use the KITTI~\cite{KITTI} dataset for training, which consists of street
traffic scenes with class labels, 2D bounding boxes and 3D annotations with
seven degrees of freedom. Our final architecture is based on \emph{Faster
R-CNN}~\cite{FasterRCNN}. The outputs of the convolutional backbone are fixed sized
feature maps for every region of interest. Fully connected layers within the
network head then propose an object class and perform 2D bounding box
regression. We extend the network head by a 3D detection head, which predicts
every degree of freedom of a 3D bounding box via classification. We achieve a
mean average precision of 47.3\% for moderately difficult data,
measured at a 3D intersection over union
threshold of 70\%, as required by the official KITTI benchmark; outperforming
previous state-of-the-art single RGB only methods by a large margin.

\end{abstract}

\section{Introduction}\label{sec:intro}

Over the past years, 2D detectors have matured and achieved excellent results.
We leverage this maturity by extending a state-of-the-art 2D detector to a 3D
detector that only relies on a single RGB image.
We use the KITTI~\cite{KITTI} dataset that consists of 3D annotated traffic scenes. 

In contrast to the four degrees of freedom of a 2D bounding box, 
a 3D bounding box has nine degrees of
freedom. These can be separated into three parts with three degrees of freedom each:
dimensions, location and orientation. We will refer to
the dimensions as height, width and length and to the location as x, y and z.
The 3D annotations of the KITTI~\cite{KITTI} dataset only provide
seven degrees of freedom because they only provide one value for the orientation.
We will refer to this orientation as angle.

The problem of 3D detection is very ill-posed when lacking depth information.
The scale, which we
will also refer to as the depth versus size ambiguity, is a common problem: An object that is far away
but big in size appears identical in an image as an object that is close to the
camera but small in size.
Many recent approaches to 3D
detection, such as \emph{PointFusion}~\cite{PointFusion},
\emph{AVOD}~\cite{AVOD}, \emph{VoxelNet}~\cite{VoxelNet} and
\emph{F-PointNet}~\cite{FrustumPointNet} use LiDAR point clouds as
their source of depth information, but we do not. 
There is also no possibility of exploiting
binocular depth cues because we only use a single RGB image. 
However, this makes our method more feasible for smaller devices, such as 
smartphones, that have neither stereo cameras nor LiDAR sensors.

In this paper we evaluate different state-of-the-art 2D detectors by debating
their suitability as a backbone for our 3D detection task. We then present
different approaches to 3D detection, including regression, classification and
human pose estimation inspired methods with
explanations about incorporating information
from the 2D detector and unobvious problems in orientation and depth prediction.
Finally, we discuss their capabilities,
problems, implementation details and detection performance.
Our evaluation shows how different subproblems contribute
to the final result and how implementation details can significantly 
impact the prediction accuracy.

For our experiments we use the \emph{Faster R-CNN}~\cite{FasterRCNN}
architecture with a \emph{ResNet-101}~\cite{ResNet} convolutional backbone
for generating feature maps, region proposals, 2D bounding boxes and object
classification. We replace the \emph{RoIPool} layer with the
\emph{RoIAlign} layer proposed in \emph{Mask R-CNN}~\cite{MaskRCNN} for
more precise local feature extraction.
\section{Related Work}

\subsection{2D Object Detection}

\subsubsection{Faster R-CNN}

The \emph{\emph{Faster R-CNN}}~\cite{FasterRCNN} architecture uses a deep
convolutional network like \emph{ResNet}~\cite{ResNet} to compute a feature
map for the entire image. A \emph{Region Proposal Network} generates regions
of interest, that are then used to create a fixed sized crop of the feature map
for every instance proposal by applying the \emph{RoIPool} layer proposed
in~\cite{FasterRCNN}. These cropped feature maps are then used to regress the
2D bounding box and classify the corresponding instance.

We use \emph{Faster R-CNN} as our baseline 2D detector. We extend the network head to allow 3D detection.

\subsubsection{Mask R-CNN}

As an extension to \emph{Faster R-CNN}~\cite{FasterRCNN}, the \emph{Mask
R-CNN}~\cite{MaskRCNN} architecture consists of the same underlying structure.
However, a feature pyramid network~\cite{FPN} is used to obtain feature maps
for various scales.
Instead of just detecting object class and 2D bounding boxes, \emph{Mask
R-CNN} addresses the task of instance segmentation, in which every object has
to be labeled with pixel-wise accuracy.

One implementation difference is the method of extracting local feature maps.
The \emph{Mask R-CNN} paper introduces a \emph{RoIAlign} layer which is used instead of the
\emph{RoIPool} layer. 

While \emph{RoIPool} performs coarse spatial quantization by rounding
coordinates of region proposals to integer values,
\emph{RoIAlign} generates the fixed size feature maps by using bilinear interpolation of the feature
map values, thus generating more precise feature maps and overall
results. According to~\cite{MaskRCNN}, this seemingly small difference has a
significant impact on their results for instance segmentation.

\emph{Mask R-CNN} uses an additional
network head to predict the instance segmentation by generating a segmentation mask for every object. This network
head consists of multiple convolutional layers and is trained with binary cross
entropy loss, where every pixel inside a region of interest is classified as
either belonging to the object or being part of the background. For
implementation details, please see~\cite{MaskRCNN}.

Our \emph{Keypoints} approach makes use of the instance segmentation
capabilities of \emph{Mask R-CNN}.

\subsubsection{YOLO9000}

Another state-of-the-art 2D detector that offers real-time performance and a
vast amount of different object classes, \emph{YOLO9000}~\cite{YOLO} claims
to outperform \emph{Faster R-CNN}~\cite{FasterRCNN} in terms of mean average
precision, while also running at 40 FPS.\@ The extremely fast inference is obtained by making use of a
one-stage network architecture: \emph{YOLO9000} is not split into a network backbone
for computing feature maps and network heads to infer object classes and bounding boxes,
but instead consists of a single convolutional network. This network, called \emph{Darknet-19},
accomplishes feature extraction, region proposal, classification and bounding
box prediction all by itself; it consists of only 19 layers, thus running
significantly faster than \emph{ResNet}~\cite{ResNet} based backbones.

Although \emph{YOLO9000}~\cite{YOLO} seems to offer more than other 2D
detectors, we eventually decided not to use their architecture due to several
reasons: Fast real-time performance
is not our main objective, this advantage is not significant for our research.
Being a one-stage detector, we expected that the underlying \emph{Darknet-19}
lacks extendibility for our purposes of 3D detection.
The network is specifically engineered for the task
of 2D detection and thus very compact and hard to modify.
It would still be possible to extract the features from the last convolutional layer and feed them to
an additional network head, but while the features generated by this
comparatively shallow network suffice for 2D detection, we are skeptical about
the feature representation being deep enough for 3D detection.

\subsubsection{RetinaNet}

Like \emph{YOLO9000}, \emph{RetinaNet}~\cite{RetinaNet} is a one-stage
detector. The paper examines why this type of detector had so far not matched
the accuracy of two-stage detectors, and concludes that the problem is
imbalance between foreground classes and the background class. To alleviate
this problem, they introduce a new loss function, called \emph{Focal Loss},
which focuses training on hard examples while reducing the importance of easy
examples that are already well-classified.

The backbone architecture of \emph{RetinaNet} is similar to that of
\emph{Faster R-CNN}, also using \emph{ResNet}~\cite{ResNet}, but
instead of a region proposal network, the features are directly fed into two
subnetworks. The first predicts a class for anchor boxes, and the second is
trained to regress from anchor boxes to ground-truth boxes.

Using this architecture, \emph{RetinaNet} surpasses existing state-of-the-art
methods in average precision, while also being faster.

Although we considered using \emph{RetinaNet} for our task, we ended up not
using it. As with \emph{YOLO9000}, we expected that \emph{RetinaNet} would
be harder to extend and adjust to our specific needs due to it being a
one-stage detector.

\subsection{3D Object Detection}

\subsubsection{F-PointNet}

Recent 3D detectors such as
\emph{PointFusion}~\cite{PointFusion}, \emph{AVOD}~\cite{AVOD},
\emph{VoxelNet}~\cite{VoxelNet} and
\emph{F-PointNet}~\cite{FrustumPointNet} rely on LiDAR data.
Currently, \emph{F-PointNet} achieves the highest 3D detection
precision; their method involves a large network that can be split into 3
different parts.

The first part is a 2D detector, such as \emph{Faster
R-CNN}~\cite{FasterRCNN}, that generates 2D region proposals and classifies the
objects. They extend these region proposals to 3D frustum proposals by
considering arbitrary depth of the object. All LiDAR points inside a
frustum are then sampled to create a point cloud for every proposal.

These point clouds are passed to the second part of the network, which is an
adaptation of \emph{PointNet}~\cite{PointNet, Pointnet++}. It carries out 3D
instance segmentation by binary classification of every point of the point cloud to
either belong to the object or background. The results of this step are
segmented object points for every frustum proposal.

The last step is estimating the amodal 3D bounding boxes.
Segmented object points are given as input to a \emph{light-weight
regression PointNet}, called \emph{T-Net}. It translates segmented
points to align their centroids with bounding box centers. Finally, 3D boxes
are estimated with another \emph{PointNet}. For details,
see~\cite{FrustumPointNet}.

We cannot adapt \emph{F-PointNet} for our task, since
we do not use LiDAR data.
Nonetheless, becoming familiar with a state-of-the-art 3D detector, that
uses LiDAR data, helped us to grasp the complexity and caveats of the task.
An example is the depth versus size ambiguity, which we have to address. It
also confirms that extending deep CNN based 2D detectors with 3D detection
utility is reasonable and produces good results.

\subsection{Depth Estimation}

For our task of 3D detection it is necessary to predict the depth of detected
objects. Although we only need a depth value for each object,
we investigated methods that perform dense monocular depth estimation
for the purpose of potentially adapting these methods.

The work of Eigen \etal~\cite{DepthEigen} is a common baseline in
this field. They use two convolutional networks: The first predicts a course
depth map of the input image. The course result is then used by the second
network to predict a refined result. While their results on the indoor
NYU~\cite{NYU} dataset are good, they point out that their method does not work
very well for KITTI~\cite{KITTI} images.
Adapting the model architecture of Eigen \etal~\cite{DepthEigen} 
for our purpose is not very feasible. Instead of two
additional convolutional networks and having to use outputs as inputs, we
would prefer to have an architecture that can directly make use of the
features computed by the 2D detector.

At this point we investigated the method of Li and He 
\etal~\cite{DepthBins1}. They use a single convolutional network to estimate
monocular dense depth of RGB images. The main emphasis of their model
architecture lies on the usage of dilated convolution~\cite{DilatedConv} and
weighted sum inference.

Dilated convolutions are used to increase the receptive field of neurons, so
that they are able to incorporate global and local depth cues. This is similar to
refining a coarse prediction, which represents global depth cues, by passing it
through another network that produces a more precise result, exploiting local
depth cues.

In~\cite{DepthBins1}, the problem of depth estimation is formulated as a
classification task. A fixed number of bins is used, each representing an
interval of depth. The depth is estimated by classifying the correct bin.
A continues output depth value is computed by using a weighted sum, i.e.\
multiplying the estimated score, or confidence, with the
corresponding bin value and then summing over all bins.

While we did not implement dilated convolutions, we ended up formulating not
just the depth estimation, but rather all parameter predictions as
classification tasks, which was inspired by Li and He 
\etal~\cite{DepthBins2}. It turned out that training our problem as a
classification task yielded better results than regression approaches.
In \fref{sec:classification}, we will discuss our usage of bin classification in the context of
3D detection and propose improvements based on the method of Li and He
\etal~\cite{DepthBins1}.

\section{Approaches}

\begin{figure}
   \includegraphics[width=1.0\linewidth]{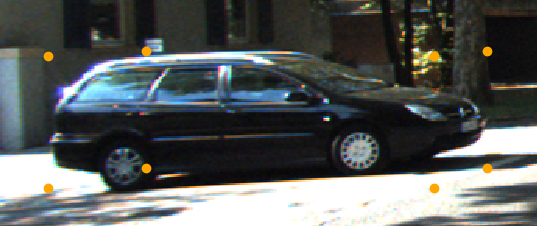}
   \caption{Corner points of a ground truth 3D bounding box.}\label{fig:keypoints}
\end{figure}

\subsection{Keypoints}
\label{sec:keypoints}

Inspired by the human pose estimation capabilities of \emph{Mask R-CNN}~\cite{MaskRCNN},
our initial approach consisted of adapting these.
Instead of predicting the position of a human shoulder or leg,
we would predict the 8 corner points of a 3D bounding box, 
treating their image coordinates as 16 degrees of freedom.

However, we soon encountered severe problems with this approach:
We assume 3D bounding boxes are cuboids, but the
result of 8 predicted pixels will most likely not be a cuboid, because there
are significantly more degrees of freedom than our desired bounding box has. If
we encode every corner point as a pair of image coordinates, 8 corner points
would already equal 16 degrees of freedom. We considered using 4 corner points,
which would result in 8 degrees of freedom, and using the geometry of a cuboid to determine the rest.
This would guarantee that the predicted shape is, if not a cuboid,
at least a parallelepiped instead of an arbitrary three-dimensional shape with 8 corner points.
But even if we assume that our network is capable of
perfectly predicting the corner points, there is still another unknown degree of freedom
to solve the depth versus size ambiguity. To obtain the final results we
would have to transform the corner points, which reside in 2D image
coordinates, to 3D camera coordinates by using the camera calibration matrix and the
predicted depth.

We realized that this approach is flawed, because it neither solves
the task of estimating the depth of an object nor does it make use of
constraints which the problem implicitly has.

\subsection{Regression}

\subsubsection{\Naive{} Regression}

\begin{figure}
        \includegraphics[width=1.0\linewidth]{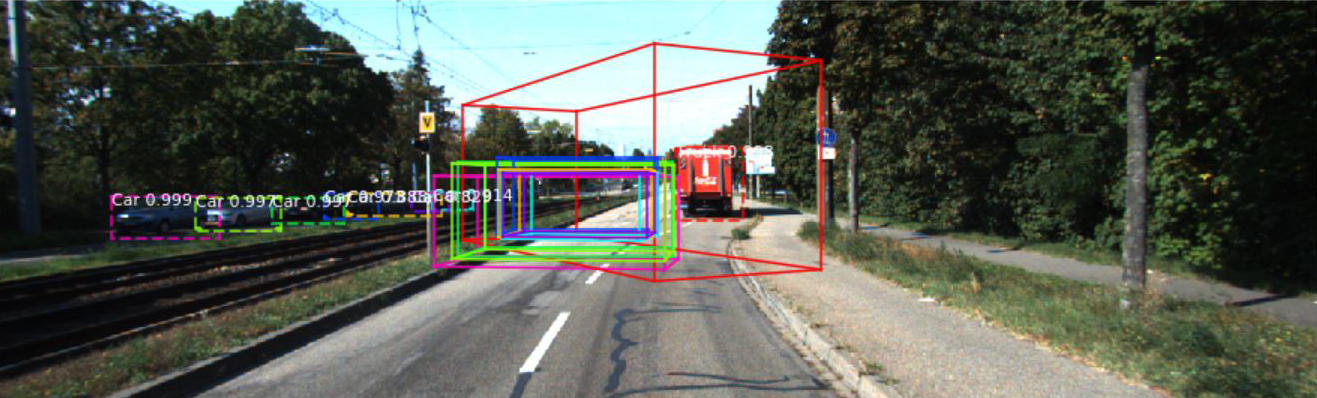}
        \caption{\Naive{} Regression confirms proposed
        problems.}\label{fig:naive}
        \includegraphics[width=1.0\linewidth]{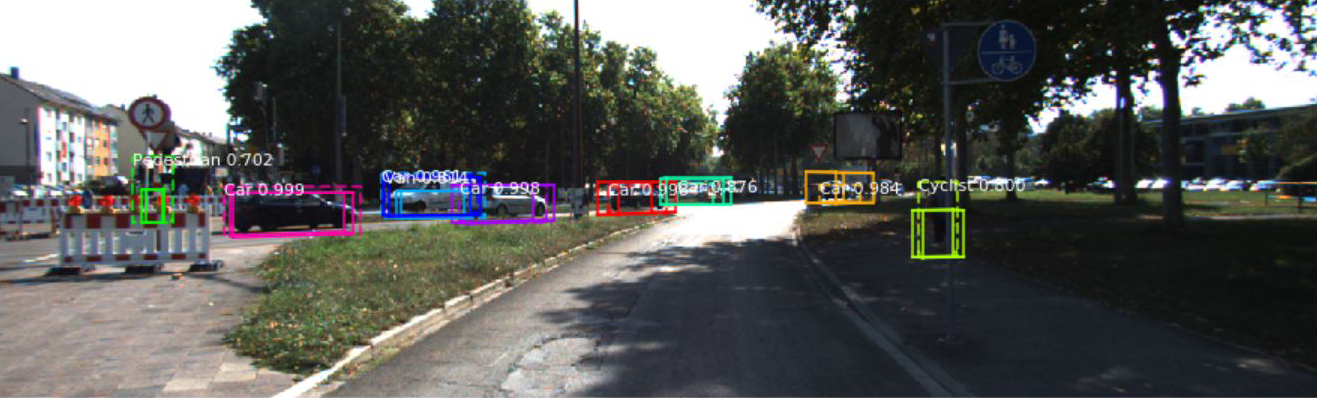}
        \caption{Our results after initial improvements.}\label{fig:regression_diff}
\end{figure}

The most \naive{} way of predicting a 3D bounding box is
using a single fully connected layer for every degree of freedom. These
layers receive a fixed sized feature map for every proposed object.
The feature maps are created by \emph{RoIAlign}~\cite{MaskRCNN} and
correspond to the regions of interest, predicted by the 2D detector.

This \naive{} method has several problems: While the network may be
able to predict the depth of an object, it cannot predict the x and y
coordinates of an object, because the local feature map is 
the only input for the layers that are
supposed to predict x and y. It does not contain any
information about where these features reside in the image. 
The network is not able to distinguish between
two occurrences of similar objects,
because their feature representations do not depend on their locations,
due to convolutional layers using the same weights at every input location.
Therefore, the network cannot
correctly predict x or y coordinates. We address this problem in the other
approaches by using information from the 2D detector. Also, training might take
very long if the network is supposed to directly output predictions for the
dimensions of objects, because it will have to learn the mean as its bias.

Although we were aware of these problems beforehand, we still decided to do the
experiments, both to confirm that our theories are correct and to
have a starting point for further improvement.

Our experiments confirmed the expected problems, see \fref{fig:naive}: 
The objects are all centered
in the x-y plane because the detector cannot predict the correct location by
only using local features.

\subsubsection{Initial Improvements}\label{sec:initial}

After our \naive{} approach indeed showed the previously proposed problems, we
constrained the regression approach by implementing prior knowledge of the data, pinhole
camera geometry and 2D detector information.

Instead of directly learning the dimensions of objects, the model learns the
difference to the mean. This change has no impact on the
potential of the neural network, because it is just an offset to the bias of
the fully connected layer. However, removing the necessity of learning
this bias yields more precise results after the same amount of training time. 

\begin{figure}
    \includegraphics[width=1.0\linewidth]{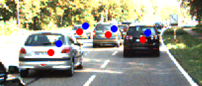}
    \caption{Comparison of the 2D bounding box center (red) and the 3D
        bounding box center, projected into image space (blue).}\label{fig:location_delta}
\end{figure}

Investigation of the ground truth data indicates that the center point of a 
2D region of interest is close to the image coordinates of the projected 3D center point (see \fref{fig:location_delta}).
We introduce a method to predict the projected 3D center of an object by
learning an offset, denoted by $du$ and $dv$, to the region of interest center predicted by the 2D detector,
denoted with image coordinates $u$ and $v$. 

Using the predicted depth value
and the camera intrinsics, we can then transform the estimated 3D center projection 
back into camera space to obtain the final 3D coordinates. 
See the supplementary material for the equations.

\subsubsection{Constrained Regression}

We attempted to implement the approach introduced by Mousavian
\etal~\cite{Constrained}. 2D bounding boxes around objects are generated
using a conventional 2D detector, which the model receives as input.
The paper assumes that these 2D bounding boxes contain the projected 3D bounding boxes,
rather than the projected objects themselves. This approximation
is accurate enough for most objects.

Under this assumption, each edge of the 2D bounding box must touch at least one
corner of the projected 3D bounding box. Since this constrains the horizontal coordinates
of two corners and the vertical coordinates of two others, it provides four
degrees of freedom.
As the KITTI dataset provides seven (cf.\ Introduction),
the network has to learn at least three more.

The degrees of freedom that are still learned are the dimensions and the angle. In our
implementation the model that predicts these is built out of linear
layers, as described in the paper. These layers receive the features produced by 
\emph{Mask R-CNN}~\cite{MaskRCNN} as input. 
This is different from the original paper, which uses features
from a pre-trained \emph{VGG} network~\cite{VGG}.

Using the 2D bounding boxes and learned degrees of freedom in combination with the camera
intrinsics, it is possible to estimate the translation of the 3D bounding box. An explanation of
how this is done can be found in the supplementary material.

The original paper explains that there is a total of 64 possibilities for which
corner points touch which edges of the 2D bounding box. The correct solution
can only be found when using the correct configuration out of these 64.
Unfortunately, the paper does not explain how the correct configuration can be
identified. Since we could not overcome this problem, we ultimately abandoned
this approach.

\subsection{Classification}\label{sec:classification}

\begin{figure}
   \includegraphics[width=1.0\linewidth]{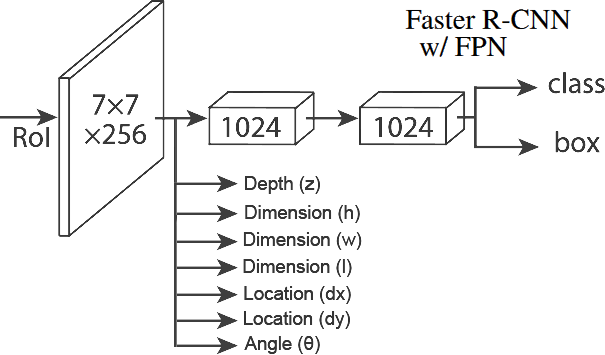}
   \caption{We extended the head architecture of Faster R-CNN. Fully connected layers that
predict the degrees of freedom of a 3D bounding box are added. 
The figure is a modified version of the figure contained in the original \textit{Mask R-CNN}~\cite{MaskRCNN} paper.}\label{fig:architecture}
\end{figure}

Due to the issues we were facing with regression, we decided to
reinterpret our task as a classification problem, because we believe these to train
better. Our network head still uses fully connected layers that
receive the local feature maps from the underlying 2D detector as input. We are
just changing the dimensions and reinterpreting the output of these fully
connected layers. The complete architecture is illustrated in \fref{fig:architecture}.

\subsubsection{Bin Representation}\label{sec:bins}

We clamped and quantized the parameters, that are to be predicted, into bins.
Every bin represents a possible range of values that a predicted parameter may
assume. Our network then learns to classify the correct bin for every
degree of freedom using the input feature map.

We choose the range for a degree of freedom by evaluating its histogram distribution,
aiming to represent more than 99\% of the training data with
our chosen clamped (and thus finite) range while also keeping it as small as
possible. Choosing a small range constrains the prediction and increases
generalizability.

After we determined the range for a degree of freedom, we chose a suitable number of
bins, depending on the variance of the training data and our estimate of the generalizability.
While intuition suggests that a higher number of bins
yields a more precise prediction, choosing a very high number of bins, i.e.\
number of classes, lowers generalizability because the range of individual bins,
i.e.\ the step size, is decreased and thus fewer training examples are available for each bin.
The goal is to choose a number of
bins that is low enough so that the network can learn to classify correctly,
but also high enough to yield a reasonably precise prediction. 
\begin{figure}
    \includegraphics[width=1.0\linewidth]{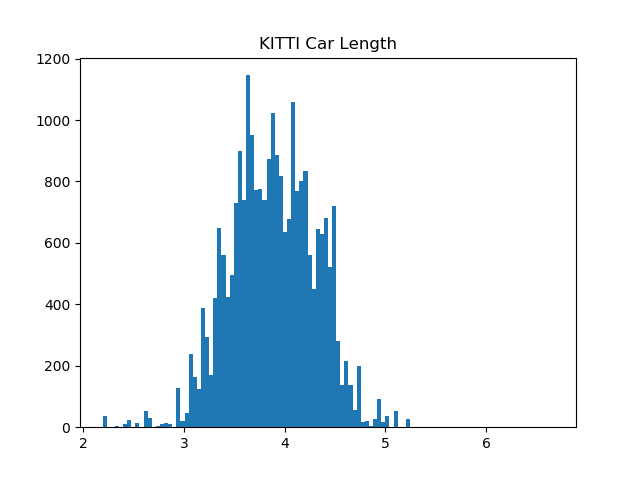}
    \caption{After evaluating this ground truth training data histogram,
        we decided to use 10 bins ranging from 3.0m to 5.0m
        with a step size of 0.2m for estimating the length of a car.}\label{fig:length}
\end{figure}

\begin{figure}
\centering
    \includegraphics[width=1.0\linewidth]{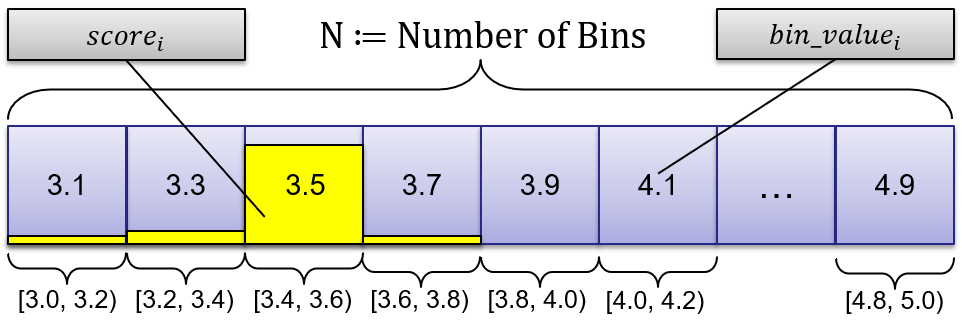}
    \caption{Visual representation of bins used for estimating length. All numbers are in meters. 
		The yellow areas represent the output of a fully connected layer.
		The \textit{lower\_bound} in this case is 3m.
		}\label{fig:bins}
\end{figure}

The final result is obtained by computing the weighted sum
over all bins.
\textit{lower\_bound} is the smallest assumable value
of the corresponding parameter.
\begin{equation*}\label{eq:bins}
    \textit{prediction} = \textit{lower\_bound} + \sum_{i = 0}^{N - 1} \textit{score}_i
    \cdot \textit{bin\_value}_i
\end{equation*}

\subsubsection{Dimensions}
We use three distinct fully connected layers to
predict the height, width and length of an object.
For the height and width of cars we use 1.2m as 
lower bound and 2.0m as upper bound with 8
equally distributed bins of step size 0.1m.
That means that e.g.\ the first bin corresponds to
the range [1.2m, 1.3m) and has bin value 1.25m. 
For the length of cars we use 3.0m and 5.0m as
lower and upper bound with 10 equally distributed bins
of step size 0.2m

\subsubsection{Depth}
After evaluating the training data, we initially
decided to use 100 bins and a range of [0m, 100m), 
with a constant step-size of
1m per bin. 

However, changing the depth of an object from 1m to 2m 
significantly impacts its appearance, while a change
from 99m to 100m may not even be noticeable in the image,
even though the absolute change is identical.
This effect happens due to perspective projection.
Estimating absolute depth is easy if an object
is close to the camera, but becomes increasingly 
difficult as the depth increases.

Hence we decided to linearly increase the step-size as
the bins represent higher depth values.
We believe that such a distribution of bins
utilizes the information in the image better
than equally distributed bins and
enables the network to predict the absolute depth
of close objects accurately. At the same time,
larger bins for high depth values increase
trainability. (cf. \fref{sec:bins})

We use 100 bins and a range of [0m, 99m). 
The first bin has step-size 0.02m, increasing linearly by 0.02m with every
bin. Our first bin covers depth values of [0, 0.02m), 
the second bin covers
[0.02m, 0.06m), the third bin covers [0.06m, 0.12m), and so on. 

\subsubsection{X and Y Location}\label{sec:xy}
Using the predicted depth and the mathematics of perspective projection, 
we can now estimate the x and y coordinates.

\begin{figure}
    \includegraphics[width=1.0\linewidth]{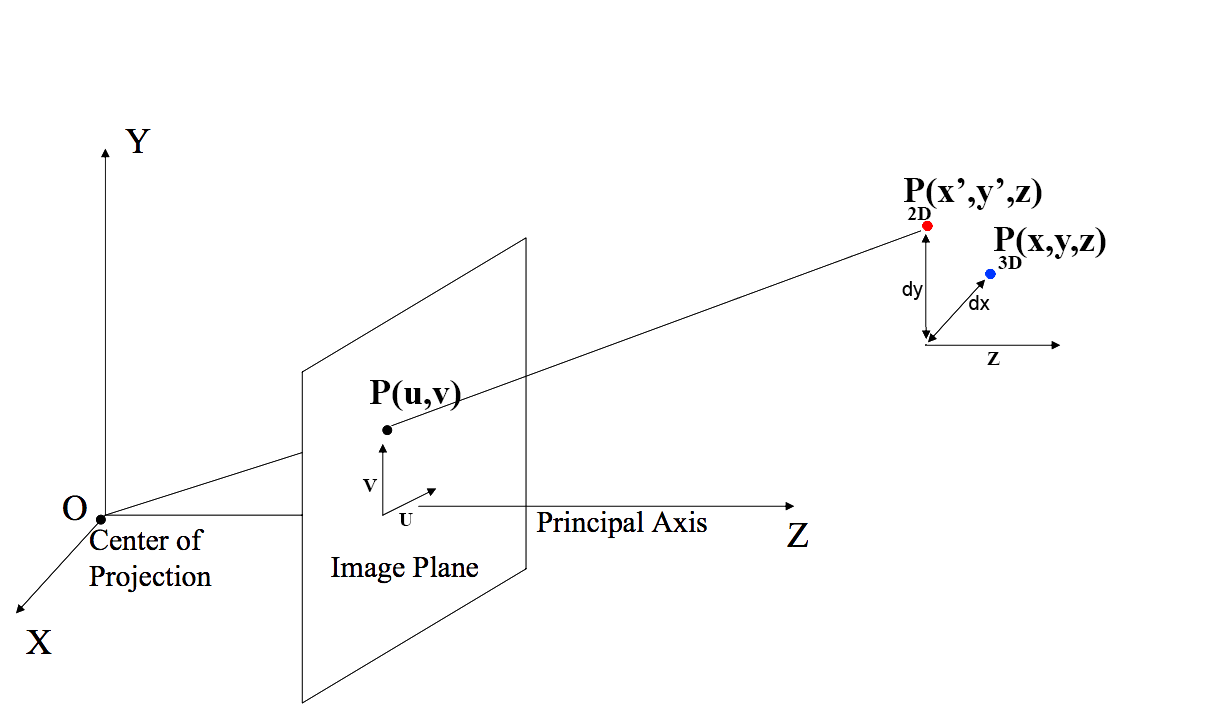}
    \caption{To obtain the 2D region of interest center in 3D camera coordinates, it is
        put at the same depth as the predicted depth of the object. The x'
        and y' coordinates are then calculated using the intrinsic camera
    matrix.}\label{fig:location_projection}
\end{figure}

Similar to how we used $du$ and $dv$ in \fref{sec:initial}, we now use offsets $dx$ and $dy$ in 3D camera coordinates.
First, we transform the 2D region of interest center (u, v) into camera coordinates (x', y', z), using the predicted depth.
$dx$ and $dy$ are the differences between this transformed region of interest center and the actual 3D object center (see \fref{fig:location_projection}).

The effect of $dx$ and $dy$ on the predicted 3D center is explained in the supplementary material.

For $dx$ we use 40 bins
with a range of [-1m,1m). 
In the case $dy$ we use 20 bins with a range of
[-0.5m,0.5m). 
Both use a
constant step-size of 0.05m.

\subsubsection{Orientation}

\begin{figure}
   \includegraphics[width=1.0\linewidth]{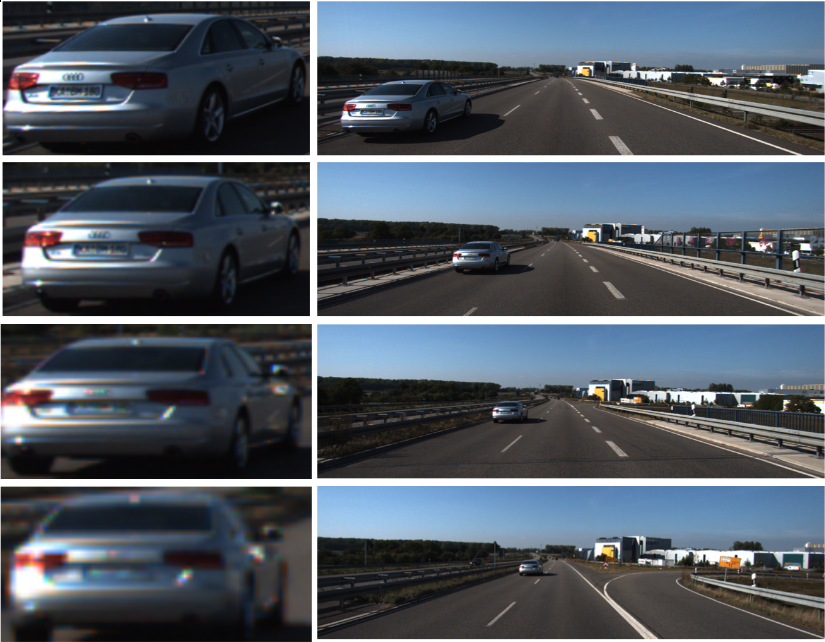}
   \caption{The car appears to rotate as it moves further from the camera.
   Source: Mousavian \etal~\cite{Constrained}}\label{fig:orientation_reason}
    \includegraphics[width=0.49\linewidth]{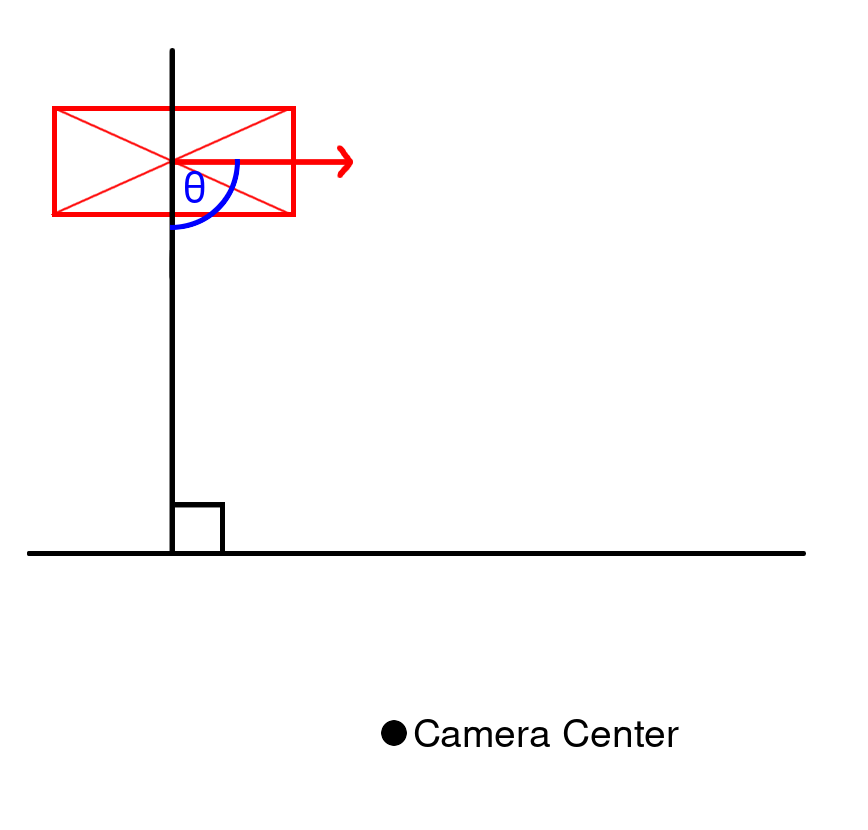}
    \includegraphics[width=0.49\linewidth]{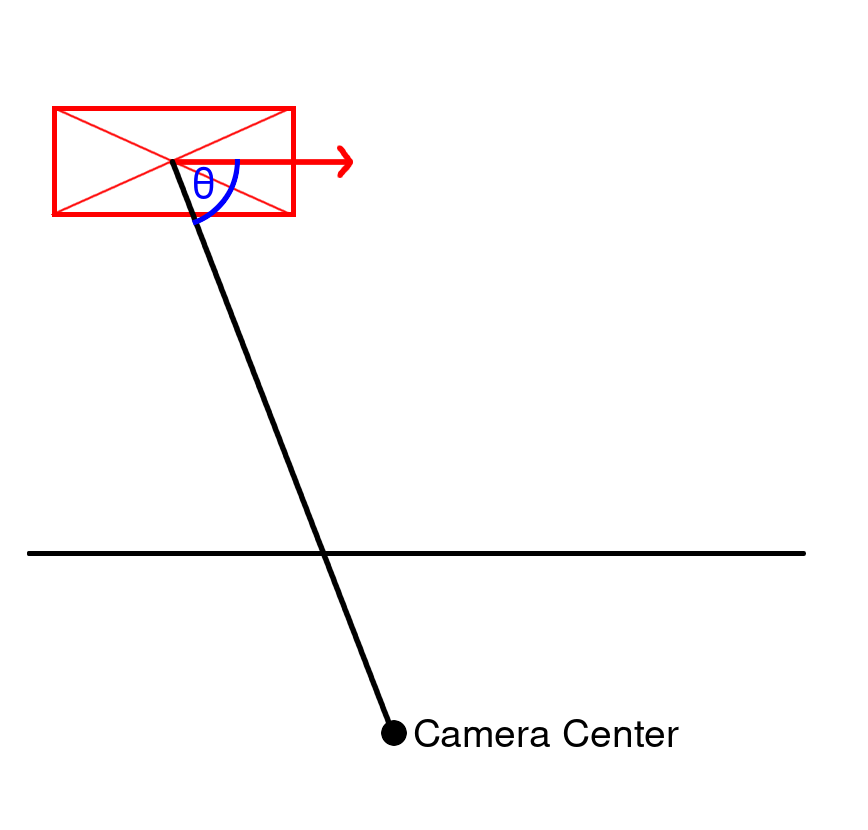}
    \caption{Instead of measuring the angle relative to a line perpendicular to
        the image plane, the angle is measured relative to a line through the
        camera center and the object center}\label{fig:orientation_fix}
\end{figure}

Since we use the KITTI dataset, we only have to predict a single angle of
orientation. We use 36 bins for this, each representing 10 degrees.

While this is just as simple as the bin representation for the dimensions,
training these bins is slightly more complex. The ground truth data has to be
pre-processed to make training easier. Why this is the case is explained
in~\cite{Constrained}, and is illustrated in~\fref{fig:orientation_reason}. If
one only observes the image patch around the car, as our model does, it looks
almost like the car is rotating while it is moving further away. Therefore, it
is easier to learn the correct orientation angle if this is taken into account.

The way this is handled can be seen in~\fref{fig:orientation_fix}. While
typically, one would use the angle with respect to a line perpendicular to the
camera plane, we instead take the angle with respect to a line connecting the
camera center and the object center.

\subsubsection{Implementation Details}

Our implementation is based on the Mask R-CNN architecture built by \textit{Matterport}~\cite{Matterport}.
This architecture contains a \textit{Feature Pyramid Network (FPN)} and a \textit{ResNet101}~\cite{ResNet} backbone.
Before the image is processed, the mean RGB values are subtracted from every pixel.
A region proposal network then creates 256 anchors using anchor scales (Length of square anchor side in pixels) of 32, 64, 128, 256, 512,
anchor ratios (width/height) of 0.5, 1, 2 and an anchor stride of 2.
The strides of each layer of the \textit{FPN} are 4, 8, 16, 32, 64.
These values are based on a \textit{ResNet101} backbone. The number of regions of interest per image that are fed to the classifier heads is set to 128.
A minimum confidence of 0.7 is required to accept a detected instance. We extend the network head architecture to include fully connected layer branches,
which predict every degree of freedom via classification. The input dimensionality of each branch corresponds to the 7x7x256 features as computed by the \textit{RoIAlign} layer.
The output dimensionality in each branch corresponds to the number of bins for every degree of freedom. A softmax activation layer is used to predict the score for each bin.
The model trains using stochastic gradient descent optimization with a learning rate of 0.002, momentum of 0.9 and weight decay regularization of 0.0001.
The losses of each degree of freedom are defined separately and penalized by a L2 regularization term. All losses contribute evenly.

\section{Evaluation}

Our evaluation is done in accordance with the official KITTI 3D object detection
benchmark evaluations, which are accessible on the KITTI
website~\cite{kittiweb}.

\begin{table}
    \centering
\begin{tabular}{r c c c}
    & Min.\ BBH & Max.\ OL & Max.\ T. \\ \hline
    Easy & 40 Px & Fully Visible & 15\% \\
    Moderate & 25 Px & Partially Occluded & 30\% \\
    Hard & 25 Px & Difficult to see & 50\%
\end{tabular}
\caption{The requirements for various difficulties according to the KITTI
    benchmark. These are taken from~\cite{kittiweb}. The
    abbreviations mean ``bounding box height'' (BBH), ``occlusion level'' (OL)
    and ``truncation'' (T).}\label{tab:kitti}
\end{table}

For this evaluation, different difficulties are defined, which specify the
objects that are taken into account. The definitions can be seen in
\fref{tab:kitti}. The models are ranked based on the results achieved with
moderate difficulty.

The best model listed is \emph{F-PointNet}~\cite{FrustumPointNet}, which achieves an average
precision of 70.39\% at moderate difficulty. However, this model relies on
point clouds from a Velodyne laser scanner, which our method does not use.
Among the models not using point clouds, the best model in the list is
\emph{3dSSD}~\cite{3dSSD}, which has an average precision of 14.97\%
at moderate difficulty. However, this method
emphasizes speed over accuracy.

\begin{table}
    \centering
\begin{tabular}{r c}
    & mAP @ 70\% IoU \\ \hline
    Easy & 55.0\% \\
    Moderate & 47.3\% \\
    Hard & 44.9\%
\end{tabular}
\caption{The mean average precisions for the difficulties defined for the KITTI
    benchmark~\cite{kittiweb}. They are measured for cars, where the
intersection over union for the 3D bounding boxes is at least 70\%.}\label{tab:diffs}
\end{table}

\begin{figure*}
	\begin{minipage}[t]{0.66\linewidth}
	\centering
      \includegraphics[width=0.49\linewidth]{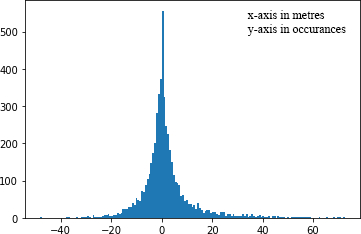}
   \includegraphics[width=0.49\linewidth]{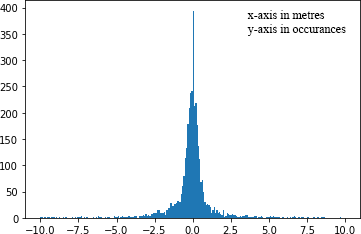}
	\begin{minipage}[t]{0.9\linewidth}
   \caption{Histograms of the difference between ground truth depth and
   predicted depth. Left: \Naive{} approach. Right: Final method. Note the x-axis scale!}\label{fig:eval_depth}
 \end{minipage}
  \end{minipage}
  \begin{minipage}[t]{0.33\linewidth}
	\centering
   \includegraphics[width=\linewidth]{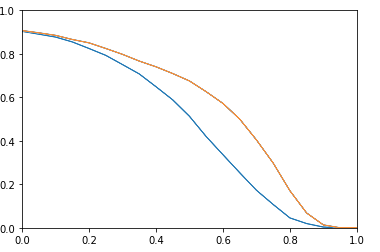}
\begin{minipage}[t]{0.9\linewidth}
   \caption{Using the central value of a bin (final model) versus using the
   smallest value of a bin as $\mathit{bin\_value}$.}\label{fig:bin_value}
  	\end{minipage}
	\end{minipage}
	\begin{minipage}[t]{0.33\linewidth}
	\centering
   \includegraphics[width=\linewidth]{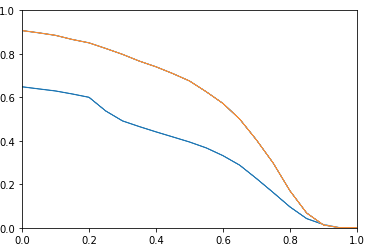}
  \begin{minipage}[t]{0.9\linewidth}
   \caption{Versus assuming
   fixed orientation.}\label{fig:ori_imp}
	\end{minipage}
  \end{minipage}
  \begin{minipage}[t]{0.33\linewidth}
\centering
   \includegraphics[width=\linewidth]{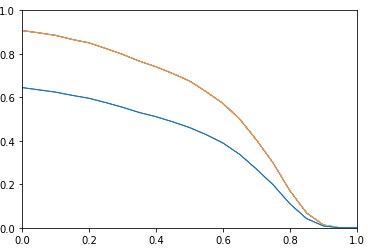}
	\begin{minipage}[t]{0.9\linewidth}
   \caption{Versus not
   taking into account $dx$ and $dy$, which were described in~\fref{sec:xy}.}\label{fig:xy_imp}
 \end{minipage}
  \end{minipage}
  \begin{minipage}[t]{0.33\linewidth}
	\centering
   \includegraphics[width=\linewidth]{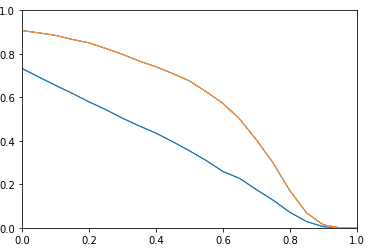}
	\begin{minipage}[t]{0.9\linewidth}
   \caption{Versus using 100
   bins with a constant step-size of 1m for depth prediction.}\label{fig:depthalt}
  	\end{minipage}
	\end{minipage}
	\begin{minipage}[t]{0.5\linewidth}
   \includegraphics[width=0.75\linewidth]{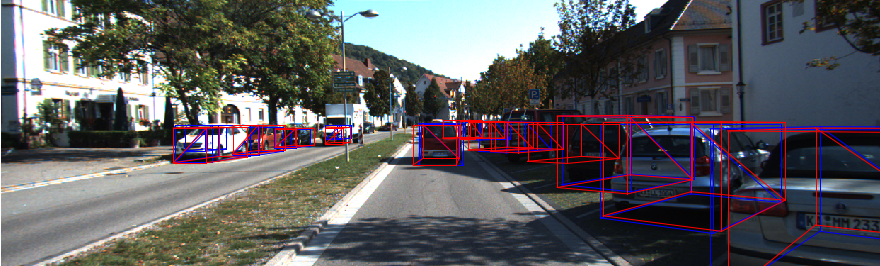}
	 \includegraphics[width=0.24\linewidth]{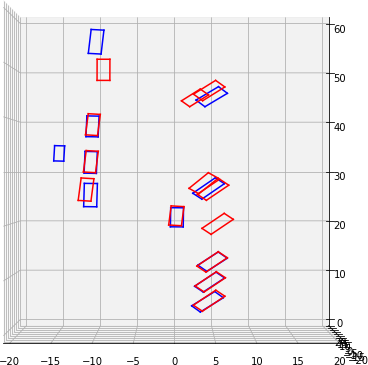}
	\end{minipage}
	\begin{minipage}[t]{0.5\linewidth}
   \includegraphics[width=0.75\linewidth]{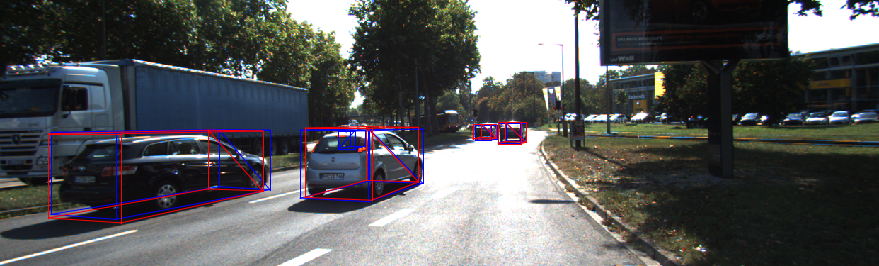}
	 \includegraphics[width=0.24\linewidth]{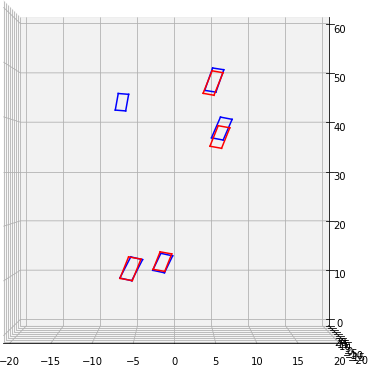}
	\end{minipage}
	\begin{minipage}[t]{0.5\linewidth}
   \includegraphics[width=0.75\linewidth]{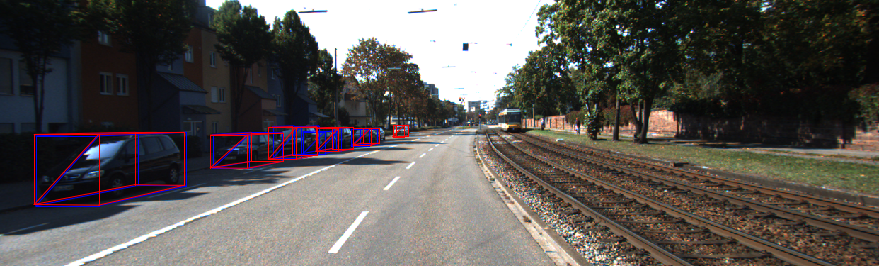}
	 \includegraphics[width=0.24\linewidth]{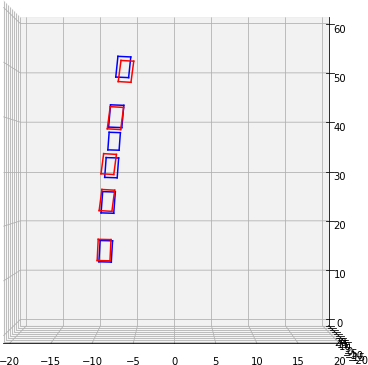}
	\end{minipage}
	\begin{minipage}[t]{0.5\linewidth}
   \includegraphics[width=0.75\linewidth]{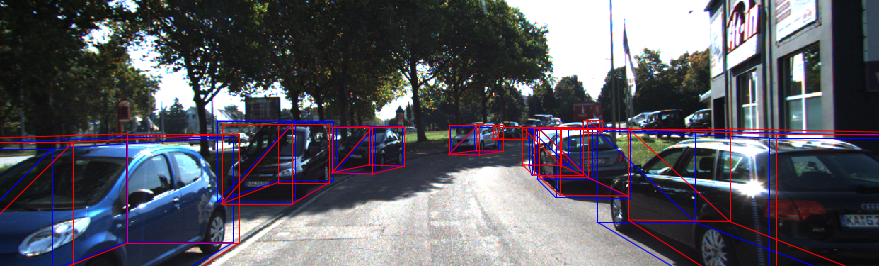}
	 \includegraphics[width=0.24\linewidth]{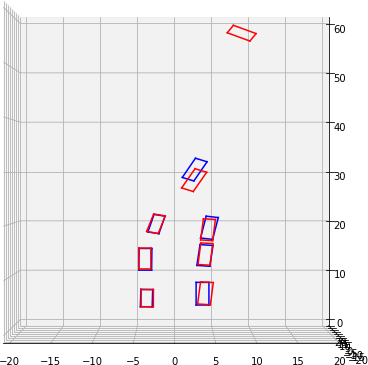}
	\end{minipage}
	\caption{Example results of our final model. The ground truth is displayed in blue, our prediction in red. On the left-hand side is the RGB image, on the right-hand side is a corresponding bird's-eye view.}
\end{figure*}

The average precision our model achieves at the various difficulties can be
seen in \fref{tab:diffs}.
The following evaluations were all made on the whole dataset, including all
difficulties. Figures \ref{fig:eval_depth} through \ref{fig:depthalt} show how our
model performs under various conditions and with various changes. The orange curve
always shows the mean average precision of the final model.

\begin{figure}
	\centering
   \includegraphics[width=0.72\linewidth]{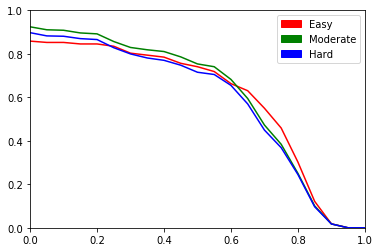}
   \caption{The average precision for various difficulties, evaluated for a
   range of IoU thresholds between 0\% and 100\%.}\label{fig:eval_all}
\end{figure}

\section{Conclusion}

Many state-of-the-art 3D detection methods use LiDAR data for their prediction.
We have shown that only using a single RGB  
image for 3D prediction can achieve high accuracy despite lacking 
important depth cues. 
Our experiments confirm that reformulating a regression problem as
a classification task can significantly improve the results. 

Our discarded human pose estimation inspired approach as described in \fref{sec:keypoints} does not properly address the problem of depth estimation.

Experiments confirm that our \Naive{} Regression cannot correctly predict the location of an object by only using local features.

Constrained Regression as introduced by Mousavian \etal~\cite{Constrained} seems very promising,
but we were not able to replicate the results.

Our Classification approach discretizes the search space into bins to predict the degrees of freedom with state-of-the-art accuracy.
Using linearly increasing bin sizes turned out to significantly improve depth estimation results.

It should be noted that our model is constrained by the chosen bin hyperparameters and
may perform worse in significantly different traffic or non-traffic scenes.
This problem could be addressed by finetuning hyperparameters and using more varied training data. 

Nonetheless, our final model successfully learned to predict the 3D layout of KITTI traffic scenes
without relying on LiDAR data or stereo cues.

\nocite{*}
\clearpage
{\small
\bibliographystyle{ieeetr}
\bibliography{egbib}
}
\vfill

\appendix
\section{Estimating Translation}
The supplementary
material to the paper about Constrained Regression [22] explains how to
obtain the translation of the 3D bounding boxes.
The following equations are provided, which have been
slightly adapted for comprehensibility:

\begin{align*}
    {\left(
        K
        \begin{bmatrix} I & R \times X_{\text{min}} \\ 0 & 1\end{bmatrix}
        \begin{bmatrix} T_x \\ T_y \\ T_z \\ 1 \end{bmatrix}
    \right)}_{\tilde{x}}
    &=
    x_{\text{min}}
    \\
    {\left(
        K
        \begin{bmatrix} I & R \times Y_{\text{min}} \\ 0 & 1\end{bmatrix}
        \begin{bmatrix} T_x \\ T_y \\ T_z \\ 1 \end{bmatrix}
    \right)}_{\tilde{y}}
    &=
    y_{\text{min}}
    \\
    {\left(
        K
        \begin{bmatrix} I & R \times X_{\text{max}} \\ 0 & 1\end{bmatrix}
        \begin{bmatrix} T_x \\ T_y \\ T_z \\ 1 \end{bmatrix}
    \right)}_{\tilde{x}}
    &=
    x_{\text{max}}
    \\
    {\left(
        K
        \begin{bmatrix} I & R \times Y_{\text{max}} \\ 0 & 1\end{bmatrix}
        \begin{bmatrix} T_x \\ T_y \\ T_z \\ 1 \end{bmatrix}
    \right)}_{\tilde{y}}
    &=
    y_{\text{max}}
\end{align*}

$K$ is the matrix containing the camera intrinsics:
\begin{equation*}\label{eq:camera_ints}
    K = \begin{bmatrix} \alpha_x & 0 & u_0 & 0 \\
        0 & \alpha_y & v_0 & 0 \\
        0 & 0 & 1 & 0 \end{bmatrix}
\end{equation*}
\(R\) refers to the rotation matrix, which can be derived from the predicted
angle. 

\(x_{\text{min}}\), \(x_{\text{max}}\), \(y_{\text{min}}\), and
\(y_{\text{max}}\) refer to the image coordinates of the horizontal and
vertical edges of the 2D bounding boxes, whereas the uppercase
version \(X_{\text{min}}\), \(X_{\text{max}}\), \(Y_{\text{min}}\), and
\(Y_{\text{max}}\) are the corners of the 3D bounding box that lie on these
edges. Note that more than one corner can lie on the same edge, in which case
the result will be the same regardless of which corner is used. 

\(T_x\),
\(T_y\), and \(T_z\) are the unknown translation parameters, which we want to find out
in order to know where the object is located in 3D space.

The \(\tilde{x}\) and \(\tilde{y}\) indices after the parentheses indicate that a coordinate
must be extracted from the homogeneous coordinate vector:

\begin{equation*}
{\left(\begin{bmatrix} x \\ y \\ z \end{bmatrix}\right)}_{\tilde{x}}
    = \frac{x}{z} \qquad
{\left(\begin{bmatrix} x \\ y \\ z \end{bmatrix}\right)}_{\tilde{y}}
    = \frac{y}{z}
\end{equation*}

The supplementary material from [22] states that it is possible to convert these into a
set of linear equations, but does not list these. As such, we calculated
them:

\begin{align*}
    \begin{bmatrix}
        -1 & 0 & {\alpha_x}^{-1}{(x_{\text{min}} - u)} \\
        0 & -1 & {\alpha_y}^{-1}{(y_{\text{min}} - v)} \\
        {-\alpha_x}{(u - x_{\text{max}})}^{-1} \hspace{-0.6cm} & 0 & -1 \\
        0 & {-\alpha_y}{(v - y_{\text{max}})}^{-1} \hspace{-0.6cm} & -1
    \end{bmatrix}
    \begin{bmatrix}
        T_x \\
        T_y \\
        T_z \\
    \end{bmatrix} \\
    =
    \begingroup
    \renewcommand*{\arraystretch}{1.2}
    \begin{bmatrix}
        r^0_x + {\alpha_x}^{-1} r^0_z (u_0 - x_{\text{min}}) \\
        r^1_y + {\alpha_y}^{-1} r^1_z (v_0 - y_{\text{min}}) \\
        r^2_z + {\alpha_x} r^2_x {(u_0 - x_{\text{max}})}^{-1} \\
        r^3_z + {\alpha_y} r^3_y {(v_0 - x_{\text{min}})}^{-1}
    \end{bmatrix}
    \endgroup
\end{align*}

where

\begingroup
\renewcommand*{\arraystretch}{1.2}
\begin{align*}
    \begin{bmatrix}
        r^0_x & r^1_x & r^2_x & r^3_x \\
        r^0_y & r^1_y & r^2_y & r^3_y \\
        r^0_z & r^1_z & r^2_z & r^3_z
    \end{bmatrix}
    =
    R \times
    \begin{bmatrix}
        X_\text{min} & Y_\text{min} & X_\text{max} & Y_\text{max}
    \end{bmatrix}
\end{align*}
\endgroup

\(T_x\), \(T_y\), and \(T_z\) can now be found by using the least squares
solution to this system of equations.

\clearpage

\section{Solving for $x$ and $y$ using offsets}

Let $K$ be the camera intrinsics matrix from \fref{eq:camera_ints}.

Let the 2D region of interest center be
$
\begin{bmatrix}
u \\ v
\end{bmatrix}
$.

Let the 3D object center be 
$
\begin{bmatrix}
x \\ y \\ z
\end{bmatrix}
$.

\subsection{Solving for $x$ and $y$ using $du$ and $dv$}
This method calculates the offset in image coordinates.

Let the projected 3D object center be
$
\begin{bmatrix}
u+du \\ v+dv
\end{bmatrix}
$.

The relationship between camera and image coordinates is described by:
\begin{align*}
z\begin{bmatrix}
u+du \\ v+dv \\ 1
\end{bmatrix}
=
K
\begin{bmatrix}
x \\ y \\ z \\ 1
\end{bmatrix}
\end{align*}

Solving for $x$ and $y$ yields:

\begin{align*}
    x &= \frac{z}{\alpha_x}\left(u + du - u_0\right) \\
    y &= \frac{z}{\alpha_y}\left(v + dv - v_0\right)
\end{align*}

\subsection{Solving for $x$ and $y$ using $dx$ and $dy$}

This method calculates the offset in camera coordinates.

Let the transformed region of interest center be
$
\begin{bmatrix}
x + dx \\ y + dy \\ z
\end{bmatrix}
$.

The relationship between camera and image coordinates is described by:
\begin{align*}
z\begin{bmatrix}
u \\ v \\ 1
\end{bmatrix}
=
K
\begin{bmatrix}
x + dx \\ y + dy \\ z \\ 1
\end{bmatrix}
\end{align*}

Solving for $x$ and $y$ yields:

\begin{align*}
    x &= \frac{z(u - u_0)}{\alpha_x} - dx \\
    y &= \frac{z(v - v_0)}{\alpha_y} - dy
\end{align*}

\end{document}